\documentclass[10pt,twocolumn]{article}

\usepackage{graphicx}
\usepackage{algorithm,algorithmic}
\usepackage{amsmath}
\usepackage{multirow}
\usepackage{float}
\newcommand\T{\rule{0pt}{2.6ex}}
\newcommand\B{\rule[-1.2ex]{0pt}{0pt}}

\begin{document}
\newcounter{cntr1}
\newcounter{cntr2}
\brokenpenalty=10000\relax

\title{Using External Archive for Improved Performance\\
in Multi-Objective Optimization
}

\author{Mahesh~B.~Patil%
\thanks{M.B.~Patil is with the Department
of Electrical Engineering, Indian Institute of Technology Bombay, Mumbai,
400076 India e-mail: mbpatil@ee.iitb.ac.in}}%

\maketitle

{\bf{This work has been submitted to the IEEE for possible
publication. Copyright may be transferred without
notice, after which this version may no longer be accessible.}}

\begin{abstract}
It is shown that the use of an external archive, purely for
storage purposes, can bring substantial benefits in multi-objective
optimization. A new scheme for archive management for the above
purpose is described. The new scheme is combined with the NSGA-II
algorithm for solving two multi-objective optimization problems,
and it is demonstrated that this combination gives significantly
improved sets of Pareto-optimal solutions. The additional
computational effort because of the external archive
is found to be insignificant when the objective functions are
expensive to evaluate.
\end{abstract}

\section{Introduction}
Anexternal archive for storing non-dominated solutions
plays an important role in a variety of multi-objective evolutionary
algorithms (MOEAs).
In algorithms based on the genetic algorithm (GA),
an external archive is generally used to store the non-dominated
solutions and propagate them to the next generation through a
suitable selection mechanism
\cite{parks}-%
\nocite{zit1}%
\nocite{paes}%
\nocite{pesa}%
\nocite{pesa2}%
\nocite{zitzler2001spea2}%
\nocite{laumanns2002}%
\nocite{fieldsend2003}%
\nocite{huang2005}%
\nocite{tiwari2008}%
\nocite{martinez2010}%
\nocite{sharma2010}%
\nocite{tiwari2011}%
\nocite{cai2015}%
\cite{song2018}.
In algorithms based on particle swarm optimization (PSO),
an external archive is used in the selection of the globally
best particle
\cite{mostaghim2003}-%
\nocite{coello2004}%
\nocite{alvarez2005}%
\nocite{tripathi2007adaptive}%
\nocite{zhang2012}%
\cite{han2017}.
In some of the evolutionary algorithms (e.g.,
\cite{pesa},\,%
\cite{tiwari2008},\,%
\cite{coello2004}),
the external archive also serves as the final output, i.e., the
Pareto-optimal solution set.

It is the purpose of this paper to show that an external archive,
even when used purely for storage, can lead to a significant
improvement in the output, viz., the set of Pareto-optimal solutions,
of an MOEA. In particular, we take {\mbox{NSGA-II}}\,\cite{deb2002},
one of the industry-standard MOEAs, as an example and present results
for some multi-objective optimization problems, with and without an
external archive, to demonstrate the advantage of using an external
archive.

The paper is organized as follows.
In Section~\ref{sec_archive_review}, we present a brief review
of the existing schemes for external archive management.
In Section~\ref{sec_new_scheme}, we discuss a new scheme
for archive management which is efficient in terms of memory requirement.
In Section~\ref{sec_new_algo}, we describe a new algorithm which is a
simple combination of the NSGA-II algorithm and the archive management
scheme of Section~\ref{sec_new_scheme}.
Finally, results obtained with the new algorithm are compared with
NSGA-II in Section~\ref{sec_results}.

\begin{figure*}
\centering
\scalebox{1.0}{\includegraphics{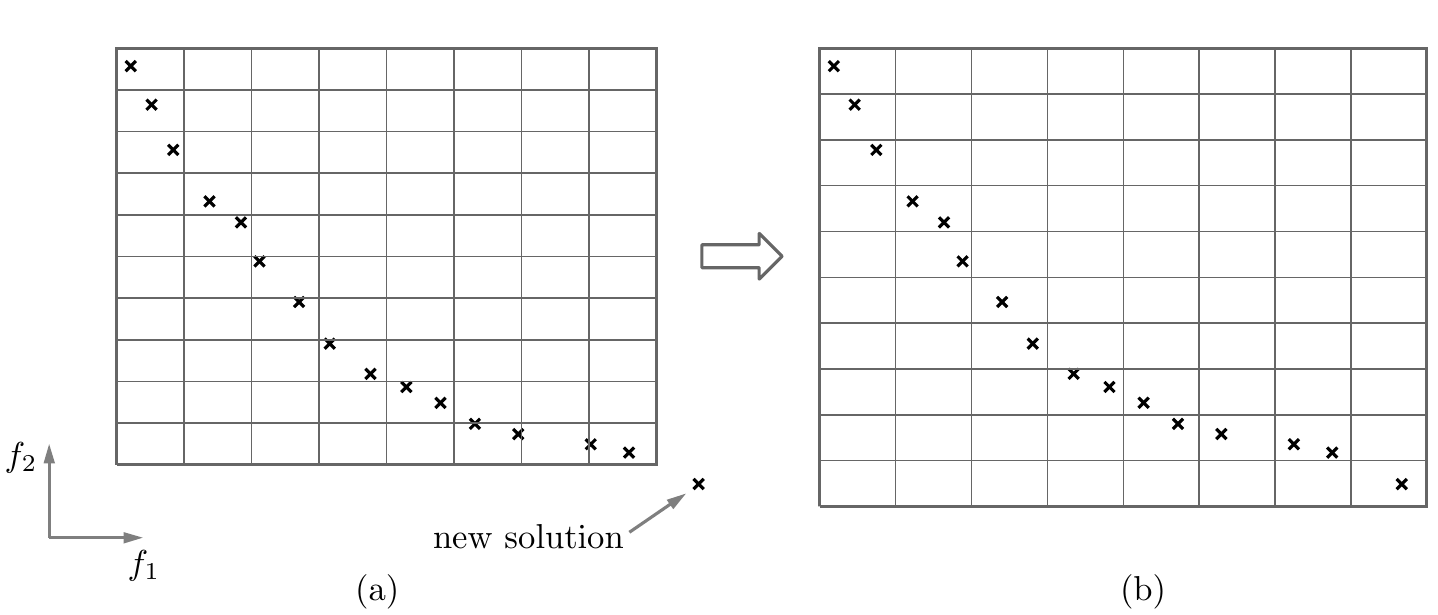}}
\caption{Schematic diagram showing the hypergrid
(a)\,before and (b)\,after a new candidate lying outside
the current hypergrid is admitted in \cite{coello2004}.}
\label{fig_coello}
\end{figure*}
\section{Archive Management: Review}
\label{sec_archive_review}
The various archiving strategies reported in the literature for
multi-objective optimization may be broadly categorized as follows.
\begin{list}{(\alph{cntr2})}{\usecounter{cntr2}}
 \item
  Unconstrained archive: In this case, the archive is not limited,
  i.e., it can hold any number of solutions
  (see \cite{parks},\,\cite{fieldsend2003},\,\cite{alvarez2005}, for
  example). This scheme has the obvious advantage that all desirable
  (non-dominated) solutions are preserved, and the efficacy of the
  search process is not compromised on account of limited archive
  size. However, the memory requirement for this approach is clearly
  larger as compared to a limited archive. Equally important, the
  archive operations are more time-consuming due to a larger number
  of archive members. Special data structures can be employed\,\cite{fieldsend2003}
  to speed up computations related to the archive.

  In a practical situation, an unlimited archive is generally not required
  from the utility perspective as long as a sufficiently large number of
  well-spread Pareto-optimal solutions are produced by the concerned MOEA.
  It is therefore more common to employ a limited archive.
 \item
  Limited archive: In this case, an upper limit ($N_A^{\mathrm{max}}$) is
  placed on the total number of solutions to be accommodated in the external
  archive. To decide whether to admit a new candidate ($C$) into the archive,
  it is compared with the existing solutions ($A_i$) in the archive. If $C$
  is dominated by each $A_i$, the archive remains unchanged. If $C$ dominates
  any of the archive solutions, they are removed from the archive and $C$ is
  admitted. In the situation where the archive is full (i.e., it has
  $N_A^{\mathrm{max}}$ solutions already) and $C$ is non-dominated with respect
  to each $A_i$, one solution needs to be removed before admitting $C$. In
  other words, the archive needs to be ``pruned" or ``truncated."

  Depending on how the non-dominated solutions are organized in the archive,
  we can have the following sub-divisions of limited archive schemes.
  \begin{list}{(\roman{cntr2})}{\usecounter{cntr2}}
   \item
    Archive with hypergrid: In this scheme, the archive is divided into an
    $M$-dimensional\,%
\cite{paes},\,%
\cite{pesa2},\,%
\cite{coello2004},\,%
\cite{han2017}
    or $L$-dimensional\,\cite{pesa}
    ``hypergrid" where
    $M$ is the number of objective functions and
    $L$ is the number of decision variables.
    The smallest unit or cell of the hypergrid is called ``hypercube"
    or ``hyperbox." For pruning of the archive, some measure of the
    density of solutions in the hypercubes is used, and a solution from a
    hypercube with a higher density is preferred for removal. In
    Section~\ref{sec_new_scheme}, we will look at the hypergrid scheme
    of \cite{coello2004} in more detail.
   \item
    Archive without hypergrid: In this case, the non-dominated solutions
    are stored as $N_A$ individual solutions and are not organized into
    hypercubes. Various approaches have been used for pruning, such as
    clustering\,%
\cite{zit1},\,%
\cite{mostaghim2003},
    distance to the closest neighbour\,%
\cite{zitzler2001spea2}, and
    crowding distance\,%
\cite{huang2005},\,%
\cite{tiwari2008},\,%
\cite{sharma2010}-%
\cite{song2018},\,%
\cite{zhang2012},
    the intention always being to remove a solution from a dense region
    of the archive.
  \end{list}
  Hypergrid-based archive management schemes are attractive from the
  computational perspective since they involve only {\it{local}}
  calculations for pruning, and in the simplest case, plain
  {\it{counting}} of solutions in a given hypercube\,\cite{coello2004}.
\end{list}

\section{Archive Management:\\New Scheme}
\label{sec_new_scheme}
In this section, we present a new scheme for archive management.
Before describing the new scheme, however, it is instructive to
look at the scheme used in the MOPSO algorithm\,\cite{coello2004}.
Fig.~\ref{fig_coello}\,(a)
shows a hypergrid example for minimisation of two objective functions
$f_1$ and $f_2$, with
$N_{f1} \,$=$\, 8$ and
$N_{f2} \,$=$\, 10$, where $N_{fk}$ is the number of divisions for the
$k^{\mathrm{th}}$ objective function.
The crosses in the figure represent the non-dominated solutions
in the archive.

If a new candidate being considered for entry into the archive
falls inside the current hypergrid~-- the outer rectangle in
Fig.~\ref{fig_coello}\,(a)~-- no changes are required in the
hypergrid boundaries. If it falls outside (see the solution
marked as ``new candidate" in the figure), the hypergrid
boundaries need to be recalculated, and the existing solutions
in the hypergrid need to be relocated, as some of them may now
fall in a different hypercube
(see Fig.~\ref{fig_coello}\,(b)).

In practice, this grid recalculation would be required more
frequently in the initial stages of the MOEA; as the algorithm
converges, the hypergrid boundaries would tend to become constant.
Nevertheless, a hypergrid scheme which does not require recalculation
of the boundaries is desirable.

Another important aspect of hypergrid management is memory
requirement. Let
$N_{\mathrm{sols}}^{\mathrm{max}}$
be the maximum number of solutions allowed in a given hypercube.
In that case, we need to allocate memory for
$N_{\mathrm{sols}}^{\mathrm{max}} N_{f1} N_{f2}$
solutions for $M \,$=$\, 2$ and for
$N_{\mathrm{sols}}^{\mathrm{max}}\times \prod_{k=1}^{M}N_{fk}$
solutions in the general case. Of these
hypercubes, only a small fraction (typically 10 to 20\,\%)
would be occupied, and in that sense, improvement in memory
utilization is desirable.

With these two improvements in mind, viz., avoiding grid boundary
recalculation and more efficient use of memory, we propose the following
hypergrid management scheme.

\begin{figure}
\centering
\includegraphics[width=0.8\columnwidth]{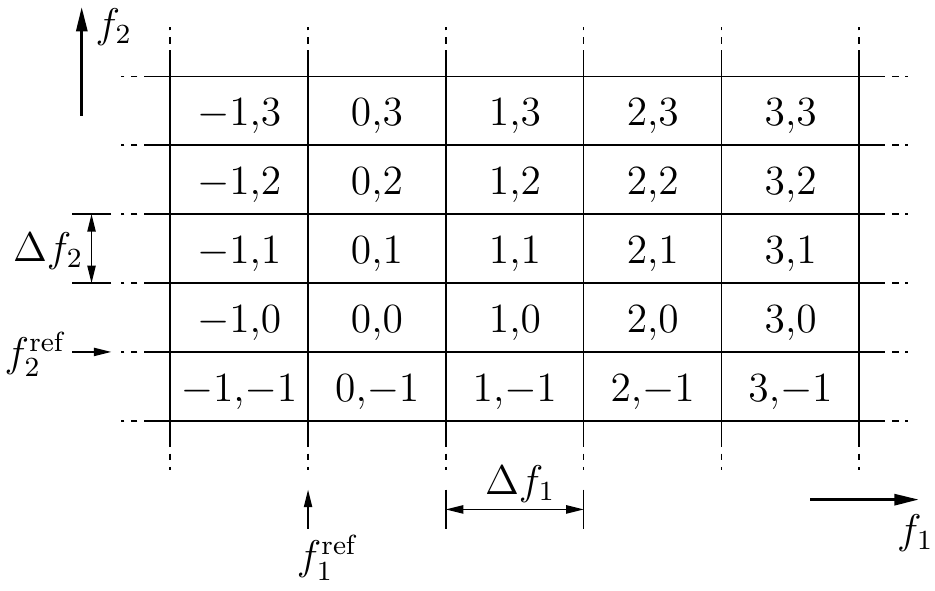}
\caption{Proposed hypergrid management scheme for
$M \,$=$\, 2$ (two objective functions).}
\label{fig_newscheme_1}
\vspace*{-3.5mm}
\end{figure}

In the new scheme, shown in
Fig.~\ref{fig_newscheme_1},
the hypergrid does not have boundaries.
Each hypercube is characterized by $M$ indices governed by the position
of the hypercube and two parameters for each of the objective functions:
a reference value $f_k^{\mathrm{ref}}$ and a spacing $\Delta f_k$.
A maximum of
$N_{\mathrm{cells}}^{\mathrm{max}}$ hypercubes (cells) are allowed, and each
hypercube can hold up to
$N_{\mathrm{sols}}^{\mathrm{max}}$ solutions. Memory allocation for
$ N_{\mathrm{cells}}^{\mathrm{max}} \times N_{\mathrm{sols}}^{\mathrm{max}} $
solutions is required in this scheme, which we will refer to as the
``fixed hypergrid" (FH) scheme.

\begin{figure*}
\centering
\scalebox{0.8}{\includegraphics{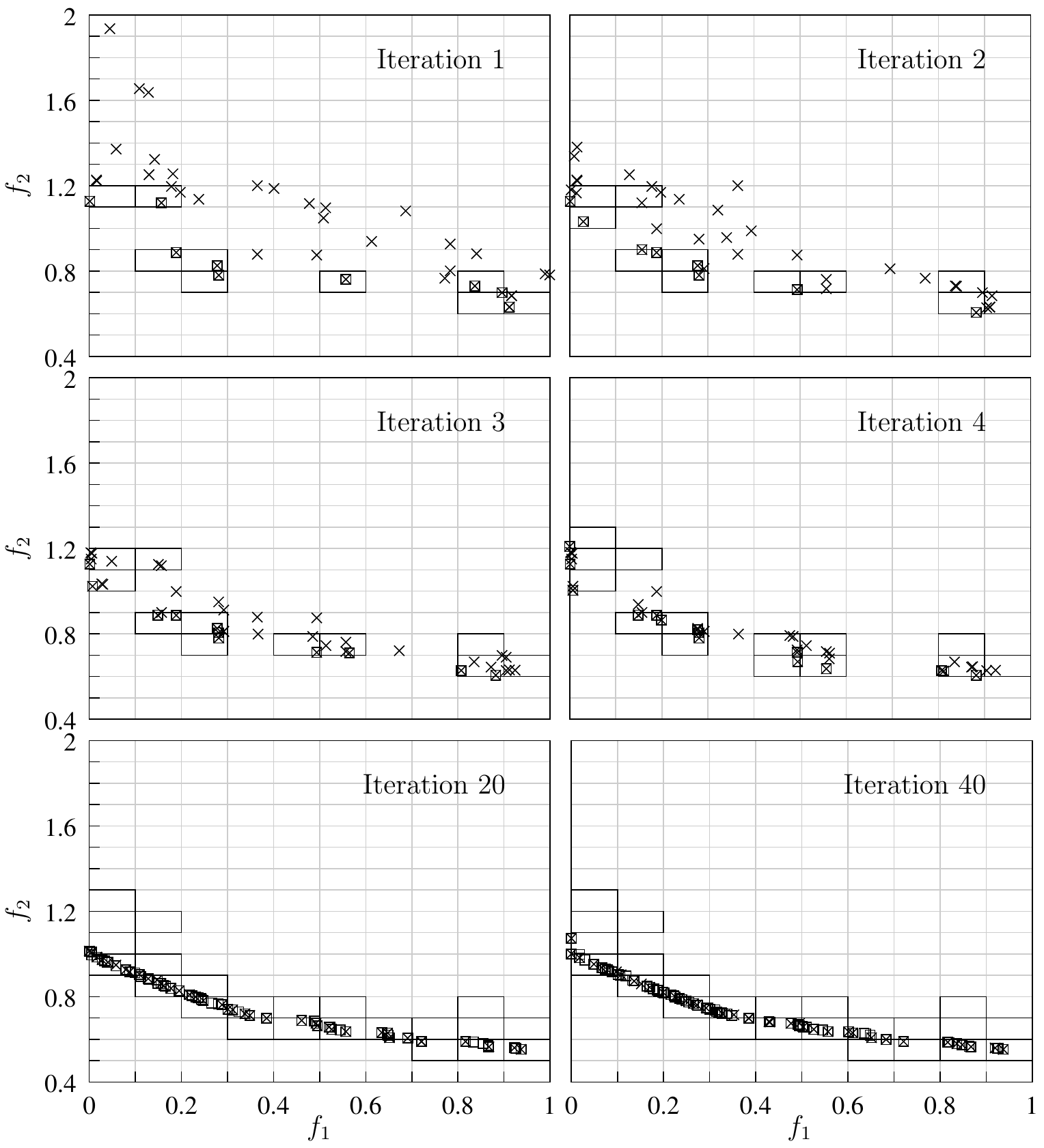}}
\caption{Illustration of the fixed grid scheme using the CTP1
two-objective problem\,\cite{deb2002}.
The real-coded NSGA-II algorithm with a population size of 
$N \,$=$\, 40$ was used. Crosses indicate the current positions
of individuals in the population, and squares indicate all of the
non-dominated solutions obtained up to the given iteration.}
\label{fig_newscheme_2}
\end{figure*}

Note that the parameter
$N_{\mathrm{cells}}^{\mathrm{max}}$ needs to be only as large as the maximum
number of occupied cells during the evolution of the population. The advantage
of the FH scheme can be seen from the results shown in
Fig.~\ref{fig_newscheme_2}
for the CTP1 two-objective problem\,\cite{deb2002}.
At the end of the first iteration, we see that 9 hypercubes (cells) have
non-dominated solutions (indicated by squares in the figure). As the
evolutionary algorithm progresses, some of the previously occupied
cells can become empty and some additional cells can become occupied
because of new non-dominated solutions appearing in the archive.
After the second iteration, four of the previously occupied cells
have become empty, two new cells have become occupied, and the total
number of cells is 11.
At the end of the $40^{\mathrm{th}}$ iteration, the number of
occupied cells is 15, and the number of empty cells
(which were occupied at some point) is 9.
We would therefore require storage for 15 cells if empty cells are
removed during the evolution and 24 if they are not.
In contrast, with the adaptive grid scheme described earlier
(Fig.~\ref{fig_coello}), we would have started with a grid of,
say $10\times 8$ (assuming roughly the same resolution as in
Fig.~\ref{fig_newscheme_2}, and kept on changing it as new solutions
joined the archive.

The FH scheme requires some parameters to be specified by the user,
viz., a reference ($f_k^{\mathrm{ref}}$) and resolution ($\Delta f_k$)
for each objective function, maximum number of cells
($ N_{\mathrm{cells}}^{\mathrm{max}}$), and maximum number of solutions
for each cell
($N_{\mathrm{sols}}^{\mathrm{max}}$).
In practice, the user may have sufficient knowledge about the optimization
problem, enabling a judicious choice for the parameters. If not, the user
can use a relatively coarse grid (large values of $\Delta f_k$), run the
MOEA, look at the results, and then fine-tune the parameters.

\begin{figure}
\centering
\includegraphics[width=0.6\columnwidth]{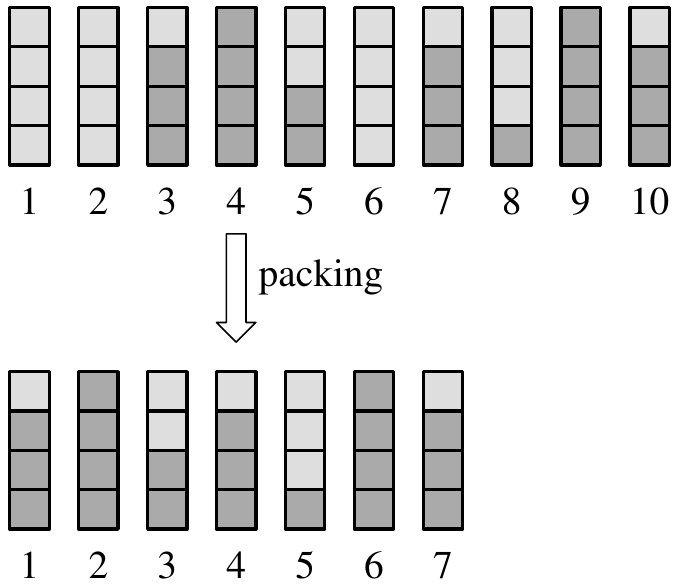}
\caption{Illustration of packing of the hypergrid, with
$ N_{\mathrm{cells}}^{\mathrm{max}} \,$=$\, 10$,
$ N_{\mathrm{sols}}^{\mathrm{max}} \,$=$\, 4$.
Occupied and empty solution slots are shown in dark grey and light
grey, respectively.
After packing, three additional cells can be added to the
archive in this example.}
\label{fig_pack}
\vspace*{-3.5mm}
\end{figure}

It is possible during the initial stages of the MOEA that a large
number of cells, which were once occupied, become empty later because of
new non-dominated solutions entering the archive. As new cells get
occupied, the total number of cells (both occupied and empty) may exceed
$ N_{\mathrm{cells}}^{\mathrm{max}}$.
When that happens, we can ``pack" the hypergrid by removing the empty cells,
as shown in
Fig.~\ref{fig_pack}.
After the packing operation, the total number of cells would
become equal to the number of occupied cells.

\begin{figure}
\centering
\scalebox{0.8}{\includegraphics{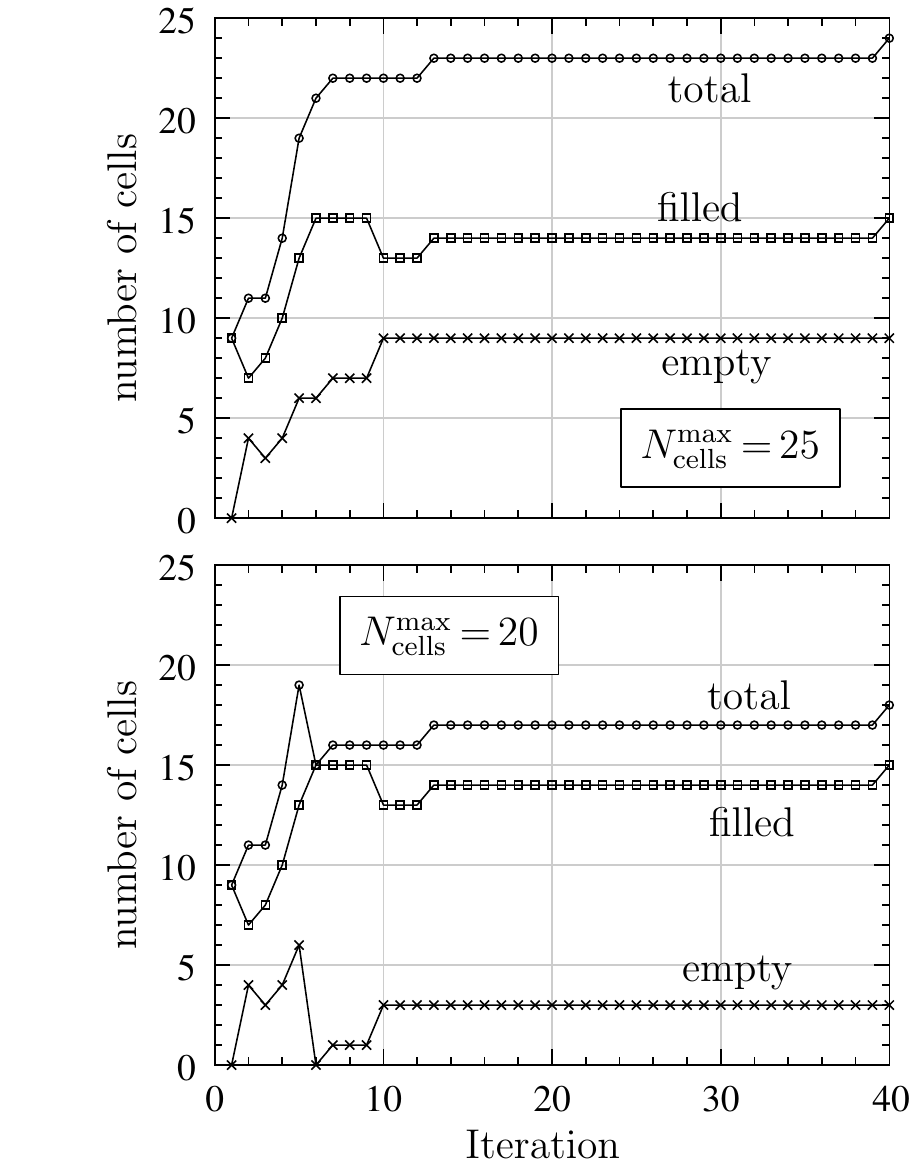}}
\caption{Filled, empty, and total number of cells versus
iteration for the CTP1 example for two values of
$ N_{\mathrm{cells}}^{\mathrm{max}}$.}
\label{fig_pack1}
\vspace*{-3.5mm}
\end{figure}

Fig.~\ref{fig_pack1} illustrates how
$N_{\mathrm{cells}}^{\mathrm{max}}$
affects the total number of cells for the CTP1 example. For
$ N_{\mathrm{cells}}^{\mathrm{max}} \,$=$\, 25$, packing does
not take place because the total number of cells is always less than
$N_{\mathrm{cells}}^{\mathrm{max}}$.
For $ N_{\mathrm{cells}}^{\mathrm{max}} \,$=$\, 20$, the total
number of cells exceeds
$N_{\mathrm{cells}}^{\mathrm{max}}$
at the 6$^{\mathrm{th}}$ iteration, and therefore a packing step
is carried out, i.e., the vacant cells are removed.

\section{NSGA-II with External Archive}
\label{sec_new_algo}
The FH scheme can be easily combined, {\it{purely}}
as a storage mechanism, with an existing MOEA. We choose NSGA-II (real-coded),
one of the industry-standard MOEAs, for this purpose. The resulting
algorithm (see Algorithm 1) will be referred to as NSGA-II-FH, i.e.,
NSGA-II with fixed hypergrid.
\begin{algorithm}
 \caption{NSGA-II-FH}
 \begin{algorithmic}[1]
  \STATE Initialize parent population.
  \STATE Initialize archive.
  \STATE Evaluate parent population.
  \STATE Assign rank and crowding distance to each individual.
  \FOR {$i_{\mathrm{gen}}$ = 1 to $N_{\mathrm{gen}}$}
   \STATE Perform selection and crossover.
   \STATE Perform mutation.
   \STATE Evaluate child population.
   \STATE Merge child and parent populations and obtain
   \STATE ~~~mixed population.
   \STATE Perform non-dominated sorting on mixed population
   \STATE ~~~and obtain the next parent population.
   \STATE Update archive.
  \ENDFOR
 \end{algorithmic}
\end{algorithm}
The only new step in this algorithm, as compared to the NSGA-II
algorithm, is step 13, that of updating the archive. Note that
the archive does not participate in the evolution of the population;
it only stores non-dominated individuals from the population in a
cumulative manner, as and when they become available.

Algorithm 2 describes the archive update procedure.
The operations involved in archive update are comparisons (to check
dominance) and copying of decision variable and objective function
values when a new solution is admitted into the archive. The FH scheme
has the advantage that it does not require recalculation of hypergrid
boundaries. However, locating $j_{\mathrm{cell}}$ corresponding to a
population member $i_{\mathrm{pop}}$ (step \ref{step1} in Algorithm~2)
is more expensive in this scheme~-- compared to the hypergrid approach
of \cite{coello2004}~-- because it involves comparing the position of
$i_{\mathrm{pop}}$ with each of the occupied hypercubes (until a match
is found). In the next section, through specific examples, we will discuss
how the additional work involved in updating the archive affects the
overall performance of the NSGA-II-FH algorithm with respect to the
standard NSGA-II algorithm.
\begin{algorithm}
 \caption{Update archive}
 \begin{algorithmic}[1]
  \FOR {$i_{\mathrm{pop}}$ = 1 to $N$}
   \FOR {$i_{\mathrm{cell}}$ = 1 to $N_{\mathrm{cells}}$}
    \FOR {each occupied solution $i_{\mathrm{sol}}$ in $i_{\mathrm{cell}}$}
      \STATE Compare $i_{\mathrm{pop}}$ and $i_{\mathrm{sol}}$ for dominance.
      \IF{$i_{\mathrm{pop}}$ and $i_{\mathrm{sol}}$ are non-dominating}
         \IF{$i_{\mathrm{pop}}$ and $i_{\mathrm{sol}}$ are identical}
           \STATE Next $i_{\mathrm{pop}}$
         \ENDIF
      \ELSIF{$i_{\mathrm{sol}}$ dominates}
         \STATE Next $i_{\mathrm{pop}}$
      \ENDIF
      \STATE  Find all solutions in archive dominated by $i_{\mathrm{pop}}$
      \STATE  ~~~and remove them.
      \STATE  Find $j_{\mathrm{cell}}$ corresponding to $i_{\mathrm{pop}}$.\label{step1}
       \IF{$j_{\mathrm{cell}}$ exists}
         \IF{$j_{\mathrm{cell}}$ is full}
           \STATE Randomly remove one solution from $j_{\mathrm{cell}}$.
         \ENDIF
         \STATE Add $i_{\mathrm{pop}}$ to $j_{\mathrm{cell}}$.
       \ELSE
         \IF{$N_{\mathrm{cells}}$ = $N_{\mathrm{cells}}^{\mathrm{max}}$}
           \IF{there are vacant cells}
             \STATE Pack archive (Remove vacant cells).
             \STATE Create a new cell and add $i_{\mathrm{pop}}$ to it.
           \ELSE
             \STATE Declare ``archive full" and stop.
           \ENDIF
         \ENDIF
       \ENDIF
    \ENDFOR
   \ENDFOR
  \ENDFOR
 \end{algorithmic}
\end{algorithm}

\section{Results and Discussion}
\label{sec_results}
Two multi-objective optimization problems are considered in this
section.
The two algorithms,
NSGA-II and
{\mbox{NSGA-II-FH}}, are used for each of the problems, keeping the
algorithmic parameters the same. In particular, the following
parameter values are used: crossover probability $p_c \,$=$\, 0.8$,
mutation probability $p_m \,$=$\, 1/L$ (where $L$ is the number
of decision variables), 
distribution index for crossover $\eta _c \,$=$\, 10$, and
distribution index for mutation $\eta _m \,$=$\, 10$.

As our first example, we consider the VNT problem (see \cite{deb2002})
which has $L \,$=$\, 2$, $M \,$=$\, 3$ (i.e., two decision variables
and three objective functions). All three functions ($f_1$, $f_2$, $f_3$)
are to be minimized, with the decision variables in the range
{\mbox{$-3 \le (x_1,x_2) \le 3$}}. We keep the population size constant
($N \,$=$\, 60$) and run NSGA-II and NSGA-II-FH for different values
of $N_{\mathrm{gen}}$ (number of generations).
The hypergrid parameters used for the NSGA-II-FH algorithm are
$N_{\mathrm{cells}}^{\mathrm{max}} \,$=$\, 1000$,
$N_{\mathrm{sols}}^{\mathrm{max}} \,$=$\, 10$,
$f_1^{\mathrm{ref}} \,$=$\, f_2^{\mathrm{ref}} \,$=$\, f_3^{\mathrm{ref}} \,$=$\, 0$,
$\Delta f_1 \,$=$\, 0.1$,
$\Delta f_2 \,$=$\, 0.01$,
$\Delta f_3 \,$=$\, 0.1$.

\begin{figure*}
\centering
\scalebox{0.8}{\includegraphics{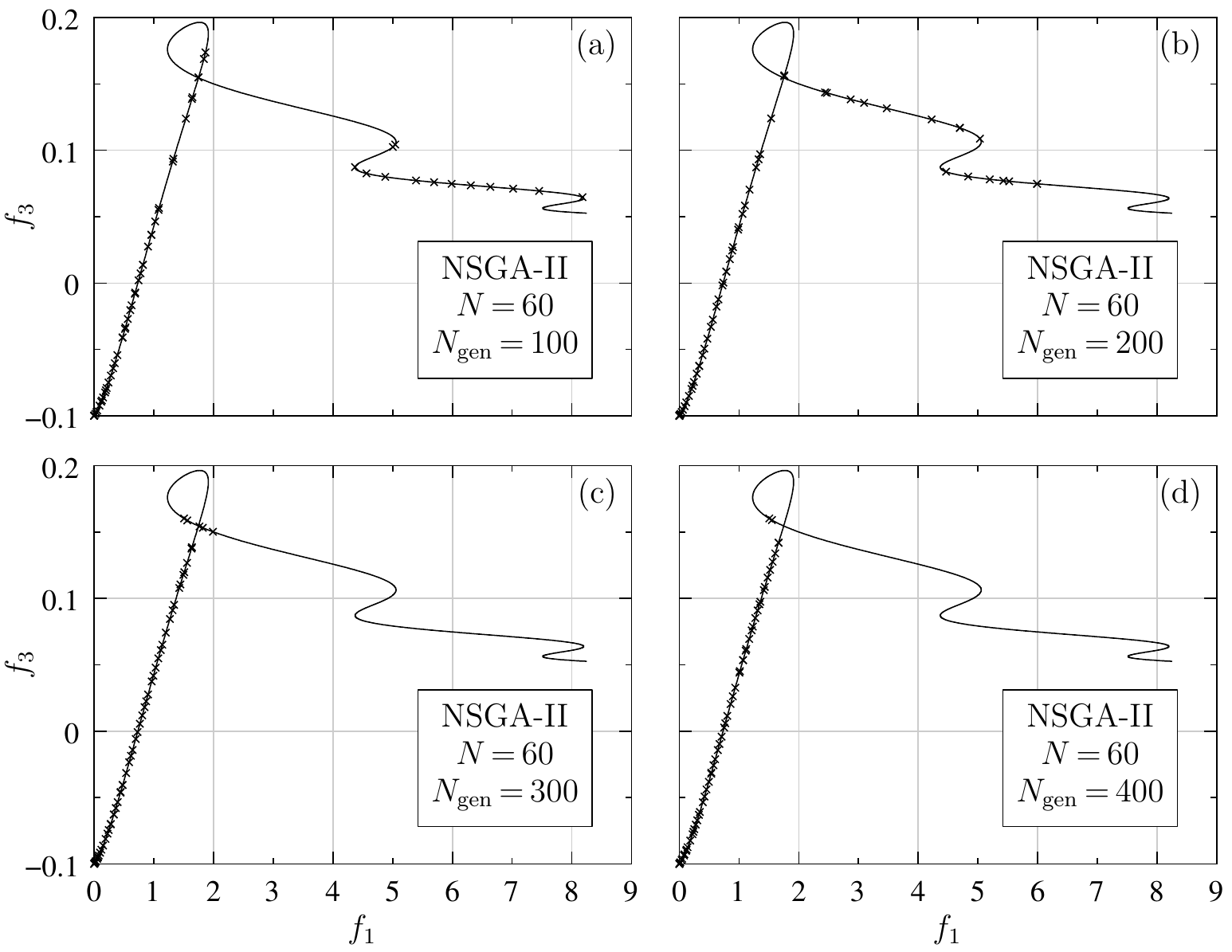}}
\caption{Pareto-optimal solutions for the VNT problem obtained with the
NSGA-II algorithm for a population size $N \,$=$\, 60$
and different values of $N_{\mathrm{gen}}$.
(a)~$N_{\mathrm{gen}} \,$=$\, 100$,
(b)~$N_{\mathrm{gen}} \,$=$\, 200$,
(c)~$N_{\mathrm{gen}} \,$=$\, 300$,
(d)~$N_{\mathrm{gen}} \,$=$\, 400$.}
\label{fig_vnt1}
\vspace*{-3.5mm}
\end{figure*}

\begin{figure*}
\centering
\scalebox{0.8}{\includegraphics{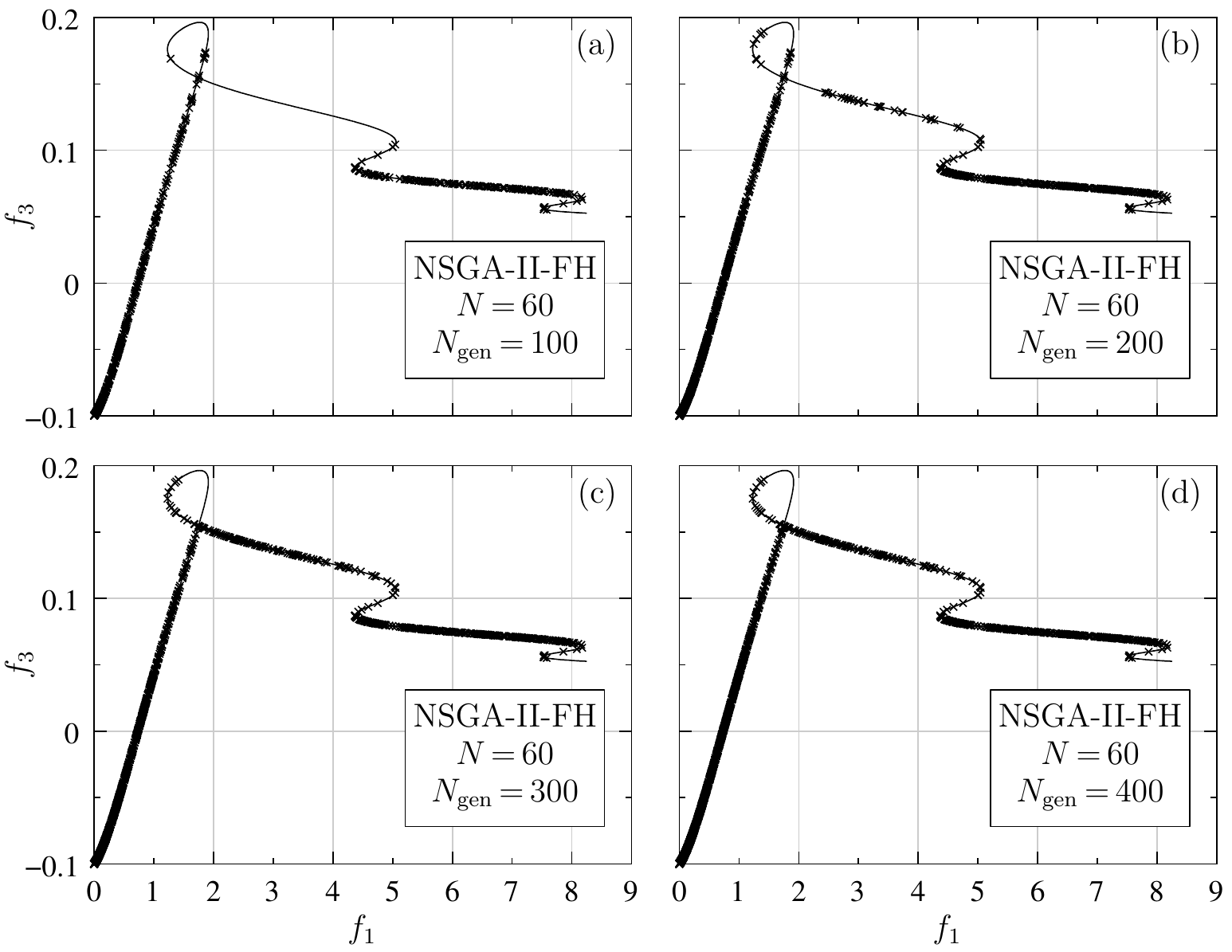}}
\caption{Pareto-optimal solutions for the VNT problem obtained with the
NSGA-II-FH algorithm for a population size $N \,$=$\, 60$
and different values of $N_{\mathrm{gen}}$.
(a)~$N_{\mathrm{gen}} \,$=$\, 100$,
(b)~$N_{\mathrm{gen}} \,$=$\, 200$,
(c)~$N_{\mathrm{gen}} \,$=$\, 300$,
(d)~$N_{\mathrm{gen}} \,$=$\, 400$.}
\label{fig_vnt2}
\vspace*{-3.5mm}
\end{figure*}

The results are shown in Figs.~\ref{fig_vnt1} and \ref{fig_vnt2}.
The true Pareto front is also shown for comparison. It can be observed
that the
NSGA-II-FH
method produces a significantly larger number of Pareto-optimal solutions
for each value of $N_{\mathrm{gen}}$.
The advantage of using an external archive to store a large number of solutions
visited by the population cumulatively is clear from the figures. This was
in fact one of the motivations behind using an external archive in the AMGA
algorithm \cite{tiwari2008}, for example. The
NSGA-II-FH
results demonstrate that an external archive is beneficial even when
it does not participate in the underlying algorithm.

\begin{figure*}
\centering
\scalebox{0.8}{\includegraphics{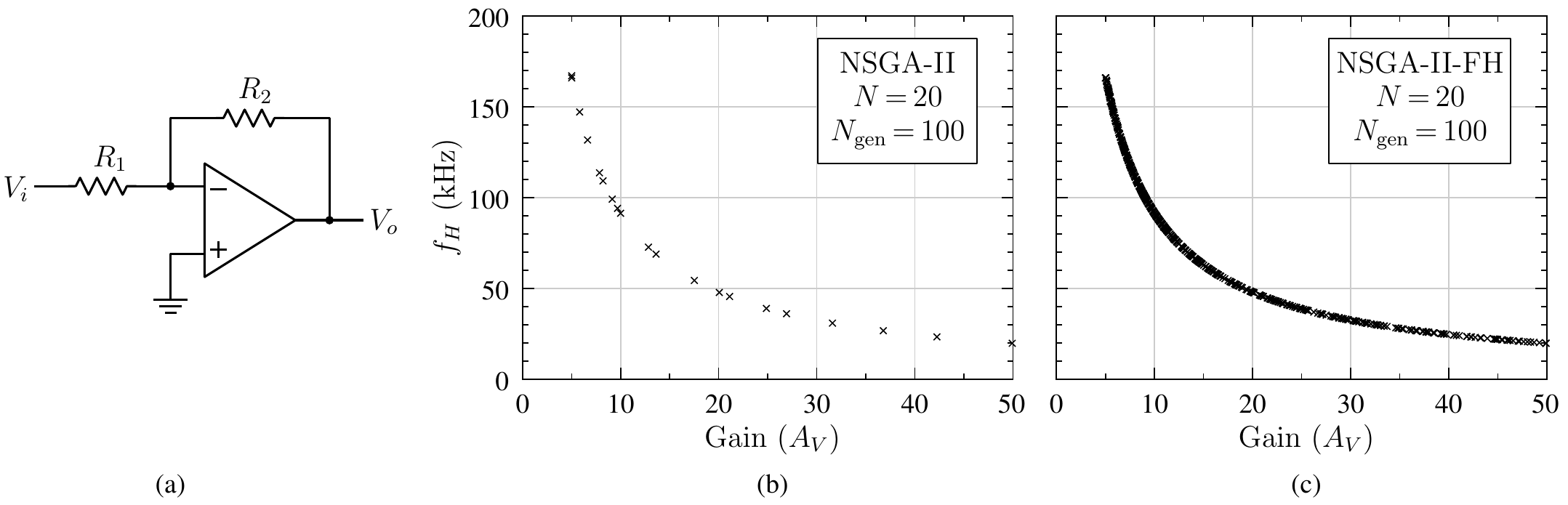}}
\caption{(a)~Op-amp based amplifier circuit, (b)~NSGA-II result,
(c)~NSGA-II-FH result obtained for a population size $N \,$=$\, 20$ and
$N_{\mathrm{gen}} \,$=$\, 100$.}
\label{fig_amp}
\end{figure*}

Our second example is an engineering problem, the inverting
amplifier shown in Fig.~\ref{fig_amp}\,(a).
It is well known that a high gain and a high bandwidth for an
amplifier are conflicting objectives. If a simple op-amp model
is used, the gain versus bandwidth relationship can be obtained
analytically. However, if a realistic op-amp model is used, circuit
simulation is required.

\begin{table*}
   \caption{CPU time taken by NSGA-II and NSGA-II-FH algorithms.}
   \centering
    \hspace*{0cm}
    \begin{tabular}{|r|r|r|r|r|}
    \hline
    {\multirow{2}{*}{$N_{\mathrm{gen}}$}}
    & \multicolumn{2}{c|}{\B \T VNT ($N \,$=$\, 60$)}
    & \multicolumn{2}{c|}{Amplifier ($N \,$=$\, 20$)}\\
    \cline{2-5}
    {} & \B \T ~NSGA-II~ & ~NSGA-II-FH~ & ~NSGA-II~ & ~NSGA-II-FH~
    \\
    \hline
    \B \T
     100\, &
     13~msec~~ &
     31~msec~~~~ &
     33~sec~~ &
     34~sec~~~~~
    \\
    \hline
    \B \T
     200\, &
     21~msec~~ &
     77~msec~~~~ &
     67~sec~~ &
     67~sec~~~~~
    \\
    \hline
    \B \T
     300\, &
     33~msec~~ &
     128~msec~~~~ &
     101~sec~~ &
     101~sec~~~~~
    \\
    \hline
    \B \T
     400\, &
     41~msec~~ &
     177~msec~~~~ &
     134~sec~~ &
     135~sec~~~~~
    \\
    \cline{1-5}
    \end{tabular}
\label{tbl_cost}
\end{table*}

The optimization problem considered here can be stated as follows.
There are two design variables, the resistance $R_1$ and $R_2$, with
{\mbox{$1\,{\rm{k}}\Omega < R_1 < 10\,{\rm{k}}\Omega$}},
{\mbox{$5\,{\rm{k}}\Omega < R_2 < 50\,{\rm{k}}\Omega$}}. There are three objective
functions: gain, bandwidth (i.e., the high cut-off frequency $f_H$), and
input resistance of the amplifier. The gain and bandwidth are to be maximised,
and the input resistance minimised. Two constraints are imposed: a minimum
gain of 5 and a minimum bandwidth of 1~kHz.

Computation of the objective functions involves setting the parameter
values ($R_1$ and $R_2$) in a circuit file, simulating the circuit using
the circuit simulator NGSPICE\,\cite{ngspice}, computing the mid-band
gain, cut-off frequency, and input resistance from the simulator output.
Note that, compared to our first example, function evaluations are much
more expensive in this case.

Each algorithm was run once with
$N \,$=$\, 20$ and $N_{\mathrm{gen}} \,$=$\, 100$.
The hypergrid parameters for the NSGA-II-FH were
$N_{\mathrm{cells}}^{\mathrm{max}} \,$=$\, 1000$,
$N_{\mathrm{sols}}^{\mathrm{max}} \,$=$\, 3$,
$f_1^{\mathrm{ref}} \,$=$\, f_2^{\mathrm{ref}} \,$=$\, f_3^{\mathrm{ref}} \,$=$\, 0$,
$\Delta f_1 \,$=$\, 1$,
$\Delta f_2 \,$=$\, 1$\,kHz,
$\Delta f_3 \,$=$\, 0.5$\,k$\Omega$.

Figs.~\ref{fig_amp}\,(b) and \ref{fig_amp}\,(c)
show the results, and it is observed once again that
NSGA-II-FH
produces a substantially larger number of Pareto-optimal
solutions.
NSGA-II, because of the absence of an external archive,
can only give up to $N$ Pareto-optimal solutions whereas
NSGA-II-FH
is not limited by the population size. In other words,
NSGA-II
gives only a sample~-- a snapshot~-- of all the Pareto-optimal
solutions visited by the population while
NSGA-II-FH
gives a more complete picture.

The benefit of the
NSGA-II-FH
algorithm comes with an additional computational cost. To gauge the
impact of this additional cost, we present in 
Table~\ref{tbl_cost}
the CPU time required for the two algorithms on a desktop computer (Linux)
with 3.3\,GHz clock and 4\,GB RAM without any parallelization.

From the table, we observe that 
NSGA-II-FH
takes substantially longer than
NSGA-II
for the VNT problem. On the other hand, for the amplifier problem,
there is virtually no difference between the two
because the time taken by function evaluations dominates in this case, and
the overhead due to archive manipulation is negligibly small.

In conclusion, a new scheme for external archive management has been
presented. It is demonstrated that the new scheme, combined with the
NSGA-II
algorithm, produces a substantially larger number of Pareto-optimal
solutions as compared to the original
NSGA-II
algorithm. The new scheme is particularly attractive for optimization
problems in which the computing time is dominated by objective function
evaluations.

\bibliographystyle{IEEEtran}
\bibliography{tevc2}

\vfill

\end{document}